\documentclass{article}
\pdfoutput=1
\usepackage[affil-it]{authblk}

\usepackage[utf8]{inputenc}
\usepackage[T1]{fontenc}
\usepackage{amssymb}
\usepackage{mathtools}
\usepackage{color, soulutf8}
\usepackage{graphicx}
\usepackage{algorithm}
\usepackage[noend]{algpseudocode}
\algnewcommand{\IfThenElse}[3]{% \IfThenElse{<if>}{<then>}{<else>}
    \State \algorithmicif\ #1\ \algorithmicthen\ #2\ \algorithmicelse\ #3
}

\usepackage{cleveref}

\newtheorem{definition}{Definition}

\begin{document}

\title{Risk Assessment of Lymph Node Metastases in Endometrial Cancer Patients: \\ A Causal Approach}
% \titlerunning{Risk Assessment of Lymph Node Metastases}

% \author{
%     Alessio Zanga \inst{1, 2, \thanks{Corresponding author: \href{mailto:a.zanga3@campus.unimib.it}{a.zanga3@campus.unimib.it} .}} \orcidID{0000-0003-4423-2121}
%     \and
%     Alice Bernasconi \inst{1, 3} \orcidID{0000-0001-8522-6882}
%     \and
%     Peter J.F. Lucas \inst{4} \orcidID{0000-0001-5454-2428}
%     \and
%     Hanny Pijnenborg \inst{5} \orcidID{0000-0002-6138-1236}
%     \and
%     Casper Reijnen \inst{5} \orcidID{0000-0001-6873-7832}
%     \and
%     Marco Scutari \inst{6} \orcidID{0000-0002-2151-7266}
%     \and
%     Fabio Stella \inst{1} \orcidID{0000-0002-1394-0507}
% }
% \authorrunning{Zanga, Bernasconi, Lucas, Pijnenborg, Reijnen, Scutari and Stella}

% \institute{
%     Department of Informatics, Systems and Communication (DISCo), University of Milano - Bicocca, Milan, Italy \and
%     Data Science and Advanced Analytics, F. Hoffmann - La Roche Ltd, Basel, Switzerland
%     \and
%     Evaluative Epidemiology Unit, Department of Research, Fondazione IRCCS Istituto Nazionale dei Tumori, Milan, Italy
%     \and
%     University of Twente, Enschede, The Netherlands
%     \and
%     RadboudUMC, Nijmegen, The Netherlands
%     \and
%     Istituto Dalle Molle di Studi sull'Intelligenza Artificiale (IDSIA), Lugano, Switzerland
% }

\author[1, 2]{Alessio Zanga}
\author[1, 3]{Alice Bernasconi}
\author[4]{Peter J.F. Lucas}
\author[5]{Hanny Pijnenborg}
\author[5]{Casper Reijnen}
\author[6]{Marco Scutari}
\author[1]{Fabio Stella}

\affil[1]{Department of Informatics, Systems and Communication (DISCo), University of Milano - Bicocca, Milan, Italy}
\affil[2]{Data Science and Advanced Analytics, F. Hoffmann - La Roche Ltd, Basel, Switzerland}
\affil[3]{Evaluative Epidemiology Unit, Department of Research, Fondazione IRCCS Istituto Nazionale dei Tumori, Milan, Italy}
\affil[4]{University of Twente, Enschede, The Netherlands}
\affil[5]{RadboudUMC, Nijmegen, The Netherlands}
\affil[6]{Istituto Dalle Molle di Studi sull'Intelligenza Artificiale (IDSIA), Lugano, Switzerland}

\maketitle

\begin{abstract}
    Assessing the pre-operative risk of lymph node metastases in endometrial cancer patients is a complex and challenging task. In principle, machine learning and deep learning models are flexible and expressive enough to capture the dynamics of clinical risk assessment. However, in this setting we are limited to observational data with quality issues, missing values, small sample size and high dimensionality: we cannot reliably learn such models from limited observational data with these sources of bias. Instead, we choose to learn a causal Bayesian network to mitigate the issues above and to leverage the prior knowledge on endometrial cancer available from clinicians and physicians. We introduce a causal discovery algorithm for causal Bayesian networks based on bootstrap resampling, as opposed to the single imputation used in related works. Moreover, we include a context variable to evaluate whether selection bias results in learning spurious associations. Finally, we discuss the strengths and limitations of our findings in light of the presence of missing data that may be missing-not-at-random, which is common in real-world clinical settings.
\end{abstract}

% \begin{keywords}
%     Causal discovery \sep
%     Causal networks \sep
%     Bayesian networks \sep
%     Missing mechanism \sep
%     Selection bias
% \end{keywords}

\section{Introduction}

\subsection{Artificial Intelligence in Medicine}

\paragraph{State of the Art.} Artificial Intelligence (AI) has found many applications in medicine \cite{Kaul2020HistoryMedicine} and, more specifically, in cancer research \cite{Troyanskaya2020ArtificialCancer.} in the form of predictive models for diagnosis \cite{Huang2020ArtificialChallenges}, prognosis \cite{Elemento2021ArtificialTherapy} and therapy planning \cite{Ho2020ArtificialTherapy.}. As a subfield of AI, Machine Learning (ML) and in particular Deep Learning (DL) has achieved significant results, especially in image processing \cite{Bi2019ArtificialApplications}. Nonetheless, ML and DL models have limited explainability \cite{Holzinger2019CausabilityMedicine} because of their black-box design, which limits their adoption in the clinical field: clinicians and physicians are reluctant to include models that are not transparent in their decision process \cite{Pumplun2021AdoptionStudy}. While recent research on Explainable AI (XAI) \cite{Gunning2019XAIExplainableIntelligence} has attacked this problem, DL models are still opaque and difficult to interpret. In contrast, in Probabilistic Graphical Models (PGMs) the interactions between different variables are encoded explicitly: the joint probability distribution $P$ of the variables of interest factorizes according to a graph $\mathcal{G}$, hence the "graphical" connotation. Bayesian Networks (BNs) \cite{Pearl2003BAYESIANNETWORKS}, which we will describe in Section \ref{sec:bns}, are an instance of PGMs that can be used as causal models. In turn, this makes them ideal to use as decision support systems and overcome the limitations of the predictions based on probabilistic associations produced by other ML models \cite{bareinboim20211OP, Lee2018StructuralIntervene}.

\subsection{Lymph Node Metastases in Endometrial Cancer Patients}

\paragraph{Background.} 
The present paper focuses on the development of a BN predictive model
for endometrial cancer (EC). Endometrial cancer is cancer of the mucous
lining, or endometrium, of the uterus. It is a common gynecological disease affecting hundreds of thousands of women worldwide. Although most patients with EC are diagnosed at an early stage of the disease and have a favorable prognosis, approximately 90,000 patients around the world die every year because of EC \cite{Bray2018}. Surgery to remove the uterus (hysterectomy), possibly together with the ovaries (ovariectomy), is the typical initial treatment for EC; the choice of neo-adjuvant (pre-surgery) or adjuvant (post-surgery) treatments depends on patient outcome prognosis. The presence of pelvic and/or para-aortic lymph node metastases (LNM) is one of the most important prognostic factors for poor outcome. The identification of LNM during the primary treatment makes it possible to choose a suitable adjuvant treatment and improve survival in node-positive EC \cite{lancet2019, Matei2019}. However, no consensus exists on how to determine which patients will benefit from lymphadenectomy (or lymph node dissection): this procedure is usually performed after or concomitant with surgery to evaluate evidence for the spread of cancer, which helps the medical team determine the progress of and treatment options for a patient's malignancy). In clinical early-stage EC, lymphadenectomy has been observed to have a marginal impact on EC outcomes and to be associated with substantial long-term comorbidities. The diagnostic accuracy for LNM is limited: approximately 50\% of LNM is found in  low- or intermediate-risk patients \cite{Bendifallah2015, Trovik2013}.

\paragraph{Objectives.} 
This work uses the BN model from Reijnen et al. \cite{Reijnen2020PreoperativeStudy} as a starting point to improve the state of the art in two ways:
\begin{itemize}
\item Extending the BN model to include the hospital of treatment as an additional variable to detect, estimate and control for potential selection bias.
\item Addressing the bias introduced by the missing imputation step, which could induce spurious correlations, hindering the interpretability of the discovered relationships.
\item Developing a causal model that integrates domain expert knowledge with observational data to better identify patients with EC designated as low or intermediate risk to develop LNM, in order to support stakeholders for decision-making.
\end{itemize}

\section{Related Work}
Individualized treatment aims to minimize unnecessary exposure to therapy-related morbidity and at the same time offers proper management according to patients' risk-stratification. In the context of EC, predicting the risk of LNM before surgical treatment has received limited attention in the literature. Koskas et al. \cite{ Koskas2016} evaluated the performance of BNs models within their cohort of 519 patients. Only one model achieved an AUC greater than 0.75,\footnote{The "Area Under the Curve", defined in page \pageref{def:auc-roc}.} highlighting the need for improved pre-operative risk stratification. Subsequent works \cite{Getz2013, Kommoss2018, VanDerPutten2016} identified biomarkers such as p53 and L1CAM as potential prognostic predictors, together with patients baseline comorbidities and tumors characteristics such as histology, grading and staging. More recently, Reijnen et al. \cite{Reijnen2020PreoperativeStudy} developed a model for the prediction of LNM and of disease-specific survival (DSS) in EC patients. This model, called ENDORISK, is a BN built on clinical, histopathological and molecular biomarkers that can be assessed pre-operatively, allowing for patient {\em counseling} and shared decision-making before surgery. ENDORISK was shown to be competitive in both goodness of fit and predictive accuracy, achieving AUC values between 0.82 and 0.85 \cite{endorisk2022}.

\section{Methods}
\label{sec:methods}

\subsection{Causal Bayesian Networks}
\label{sec:bns}

Firstly, we will summarize those key definitions for BNs and causal models that we will need to describe our contributions in Section \ref{sec:methods}.

\begin{definition}[Graph]
    A graph $\mathcal{G} = $(\textbf{V}, \textbf{E}) is a mathematical object represented by a tuple of two sets: a finite set of nodes \textbf{V} and a finite set of edges $\textbf{E} \subseteq \textbf{V} \times \textbf{V}$. In the following pages (\textbf{V}, \textbf{E}) will be omitted if not specified otherwise.
\end{definition}

We will focus on \textit{directed graphs} where $(X, Y) \neq (Y, X)$, which is graphically represented as $X \rightarrow Y$. A directed graph encodes a set of ordinal relationships, i.e. in $X \rightarrow Y$ the node $X$ is called \textit{parent} of $Y$ and $Y$ is said to be the \textit{child} of $X$. Therefore, the set of parents of $X$ is $\mathit{Pa}(X)$, while the set of children of $X$ is $\mathit{Ch}(X)$. \\

A directed path $\pi$ is a finite ordered set of nodes $\pi = (V_0 \rightarrow \dots \rightarrow V_n)$ such that each adjacent pair of nodes $(V_i, V_{i+1})$ in $\pi$ is a directed edge in $\textbf{E}$. A cycle is a path where the first and the last node are the same node. A graph is acyclic if it contains no cycle, also called a Directed Acyclic Graph (DAG).

\begin{definition}[Causal Graph]
    A causal graph $\mathcal{G} = (\mathbf{V}, \mathbf{E})$ {\rm \cite{bareinboim20211OP}} is a graph that encodes the cause-effect relationships of a system.
\end{definition}

\paragraph{Causes \& Effects.} The set $\mathbf{V}$ contains the variables that describe the behavior of the system under study, whereas the set $\mathbf{E}$ contains the edges that make explicit the interplay of the variables. In particular, for each directed edge $(X, Y) \in \mathbf{E}$, $X$ is said to be a direct cause of $Y$, whereas $Y$ is called direct effect of $X$. This definition is recursive: a variable $Z$ that is the direct cause of $X$, but not of $Y$, is said to be an indirect cause of $Y$. \\

This mapping between a causal graph $\mathcal{G}$ and the cause-effect relationships is formalized by the causal edge assumption \cite{pearl2016causal}.

\begin{definition}[Causal Edge Assumption]\label{def:causal_edge} Let $\mathcal{G} = (\mathbf{V}, \mathbf{E})$ be a causal graph. The value assigned to each variable $X \in \mathbf{V}$ is completely determined by the function $f$ given its parents:
\begin{equation}
    X \coloneqq f(\mathit{Pa}(X)) \qquad \forall X \in \mathbf{V}
\end{equation}
\end{definition}

The causal edge assumption allows us to interpret the edges of a causal graph in a non-ambiguous way: it enforces a recursive relationship over the structure of the graph, establishing a chain of functional dependencies. Hence, this class of graphical models is inherently explainable, even for researchers approaching them for the first time. \\

When the causal graph is not known a priori, it is possible to recover it from a combination of prior knowledge and data driven approaches. Such problem is called Causal Discovery \cite{Zanga2022APracticeb}.

\begin{definition}[Causal Discovery]
    Let $\mathcal{G}^*$ be the true but unknown graph in the space of possible graphs $\mathbb{G}$ from which the data set $\mathbf{D}$ has been generated. The Causal Discovery problem consists in recovering $\mathcal{G}^*$ given the data set $\mathbf{D}$ and the prior knowledge $\mathbf{K}$.
\end{definition}

Once the causal graph $\mathcal{G}^*$ is recovered, it is possible to build a PGM with the given structure. For example, BNs \cite{Pearl2003BAYESIANNETWORKS} are a widely known type of PGM.

\begin{definition}[Bayesian Network]
    Let be $\mathcal{G}$ a DAG and let $P(\mathbf{X})$ be a global probability distribution with parameters $\Theta$. A BN $\mathcal{B} = (\mathcal{G}, \Theta)$ is a model in which each variable of $\mathbf{X}$ is a vertex of $\mathcal{G}$ and $P(\mathbf{X})$ factorizes into local probability distributions according to $\mathcal{G}$:
    \begin{equation}
        P(\mathbf{X}) = \prod_{X \in \mathbf{X}} P(X \, | \, \mathit{Pa}(X))
    \end{equation}
\end{definition}

The key difference between a BN and a Causal BN (CBN) is the semantic interpretation of its edges. Indeed, in a CBN an edge represents a cause-effect relationship between two variables, whereas the same edge in a BN entails only a probabilistic dependence.

\begin{definition}[Causal Bayesian Network]
    A Causal BN $\mathcal{B} = (\mathcal{G}, \Theta)$ is a BN where the associated DAG $\mathcal{G}$ is a causal graph.
\end{definition}

\subsection{Causal Discovery with Observational and Missing Data}
Causal discovery algorithms are usually divided into two classes: constraint-based and score-based. The two classes have been extended to handle  missing data in different ways: constraint-based algorithms rely on test-wise deletion \cite{Strobl2018FastDeletion} to perform conditional independence tests efficiently in order to mitigate the impact of missing observations, while score-based approaches alternate data imputation and causal discovery \cite{Friedman1998TheEM}.

\paragraph{Causal Discovery with Missing Data.} By default, causal discovery algorithms are not designed to handle incomplete data. However, we can combine them with missing value imputation approaches to complete the data and reduce the problem to a standard causal discovery. A widely-used application of this idea is the Expectation Maximization (EM) \cite{Lauritzen1995TheData} algorithm. In particular, the Structural EM \cite{Friedman1998TheEM} algorithm is specifically designed to iteratively run the imputation step performed by EM and a causal discovery step performed by a score-based algorithm, alternating them until convergence.

\paragraph{Greedy Search: The Hill-Climbing Approach.} A widely applied score-based algorithm for causal discovery is Greedy Search (GS) \cite{Scutari2019LearningImplementation}. GS traverses the space $\mathbb{G}$ of the possible DAGs over the set of variables $\mathbf{V}$, selecting the optimal graph $\mathcal{G}^*$ by a greedy evaluation of a function $\mathcal{S}$, known as the \textit{scoring criterion}. There are multiple strategies to implement GS, one of which is called Hill-Climbing (HC). At its core, HC repeatedly applies three fundamental operations to change the current recovered structure, moving from a graph to another, across the graphs space $\mathbb{G}$. These ``moves'' are the addition, deletion or reversal of an edge. If a move improves the score $\mathcal{S}$, then the graph is updated accordingly. The procedure halts when no moves improve the score and returns a DAG. \\

While the graphs space $\mathbb{G}$ contains every graph that could be generated given the vertices $\mathbf{V}$, only a subset of them are compatible with the probability distribution induced by the observed data. Moreover, not every graph compatible with said distribution is necessarily causal. Therefore, it is possible to shrink the search space by adding constraints in terms of structural properties, that is, by requiring or forbidding the existence of an edge in the optimal graph $\mathcal{G}^*$.

\paragraph{Encoding Prior Knowledge.} One could restrict the set of admissible graphs by encoding prior knowledge through required or forbidden edge lists \cite{Meek2013StrongNetworks}. For instance, it is possible to leverage expert knowledge to identify known relationships and encode them as required edges. These lists can also encode a partial ordering when potential causes of other variables are known. \\

For example, suppose that clinicians want to include their prior knowledge on the interaction between biomarkers and LNM into the CBN. This inclusion would happen during the execution of the causal discovery algorithm and, therefore, requires that the experts' knowledge is encoded programmatically. Causal discovery algorithms essentially learn a set of ordinal, parent-child relationships: it is natural to encode prior knowledge in the same form. For instance, if we know that p53 is not a direct cause of LNM, then the translation of such a concept would be p53 $\not \in \mathit{Pa}($LNM$)$. If, on the other hand, we know that LNM is a direct cause of L1CAM then we would have L1CAM $\in \mathit{Pa}($LNM$)$ . This is a direct consequence of the Causal Edge Assumption (\Cref{def:causal_edge}). Even this simple example shows the flexibility of this approach, allowing to encode different sources of prior knowledge without any restrictions.

\section{Experimental Results}

Causal discovery algorithms provide a correct solution to the causal discovery problem in the limit of the number of samples \cite{spirtes2000causation}. However, in real-world applications the available data are finite, especially in medicine, where data samples are usually small. As a result, even small amounts of noise in the data may result in a different structure. Therefore, it is important to quantify our confidence in the presence of each edge in the causal BN, also called the "strength" of an edge.

\paragraph{Estimating Edge Strength: A Bootstrap Approach.} The estimation of the strength of an edge was performed through a bootstrap approach \cite{Friedman2013DataApproach}. Here, a custom version with Structural EM is reported in \Cref{alg:confidence}, described as follows. Line 1, the procedure takes as input a data set $\mathbf{D}$, prior knowledge $\mathbf{K}$, hyperparameters $\pmb{\alpha}$ for Structural EM, number of bootstraps $n$ and number of samples to draw $m$. Line 2, the confidence matrix $\mathbf{C}$ is initialized. Lines 3-6, the data set $\mathbf{D}$ is re-sampled $n$ times with replacement, drawing $m$ observations for each bootstrap following a uniform distribution. For each sampled data set $\mathbf{D}_i \subseteq \mathbf{D}$, the causal discovery algorithm is applied to induce a corresponding graph $\mathcal{G}_i$. Finally, line 7, is responsible to compute the strength of each edge as the relative frequency of inclusion across the $n$ bootstraps. \\

The causal discovery algorithm developed is described in \Cref{alg:learncbn}. Line 1 is based on the confidence matrix estimation computed by \Cref{alg:confidence}. Line 2, the causal graph $\mathcal{G}$ is initialized to the empty graph and, line 3, the associated confidence matrix $\mathbf{C}$, i.e. the matrix containing the edges strength, is computed. Line 4 describes a generic strategy to select the edges to insert into $\mathcal{G}$ given $\mathbf{C}$. Here, we relied on a threshold $\lambda$ to filter irrelevant edges to build the ``average graph''. Lines 5-6, the CBN parameters $\Theta$ are fitted given $\mathcal{G}$ by applying EM \cite{Lauritzen1995TheData} to the data set $\mathbf{D}$ with missing data.

\paragraph{Definition and Selection of Variables.} To conduct this analysis we used the cohort presented by Reijnen et al. An overview of the cohort and the procedures done for data collection can be found in \cite{Reijnen2020PreoperativeStudy}. Briefly, the retrospective multicenter cohort study included 763 patients, with a median
age 65 years, surgically treated for endometrial cancer between 1995 and 2013 at one of the 10 participating European hospitals. Clinical and histopathological variables with prognostic value for the prediction of LNM were identified by a systematic review of the literature. The used variables could be divided into three major temporal tiers: 
\begin{itemize}
\item \textit{Pre-operative clinical, histopathological variables and biomarkers}: Estrogen Receptor (ER) expression, Progesteron Recepter (PR) expression, L1CAM (cell migration) expression, p53 (tumour suppressor gene) expression, cervical cytology, platelets counts (thrombocytosis), lymphadenopathy on MRI or CT, lymphovascular space invasion (LVSI), Ca-125 serum levels and pre-operative tumor grade,
\item \textit{Post-operative/treatment variables:} adjuvant therapy (Chemotherapy and/or Radiotherapy), post-operative tumour grade,
\item \textit{Late post-operative outcomes:} 1-,3-,5-year disease-specific survival (DSS), Lymph Nodes Metastases (LNM), Myometrial Invasion.
\end{itemize}
All the described variables are discrete variables, with cardinality ranging from 2 to 3. Two main changes were done in comparison to published works: addition of hospital of treatment (10 levels) in the model and separation of adjuvant therapy into two different dichotomous variables (chemotherapy and radiotherapy).

\paragraph{Training and Testing.} The data set $\mathbf{D}$ was split in a train set and a test set following a 70/30 ratio. For each configuration of hyperparameters $(\pmb{\alpha}, n, m, \lambda)$, we applied \Cref{alg:learncbn} to the train set, with the same prior knowledge $\mathbf{K}$. The resulting BNs were evaluated on the test set by estimating the probability of LNM. The hyperparameter tuning was performed following a grid search, as suggested in \cite{spirtes2000causation}. While \emph{cross validation} (CV) is generally preferred over a na{\"i}ve train-test splitting, hyperparameter tuning over a learning procedure based on \emph{Structural EM} is computationally expensive and, therefore, it would require a nonignorable amount of time when coupled with CV. Moreover, we considered the possibility to further split the train set to obtain a \emph{validation set}, but the reduced sample size hindered the feasibility of this additional step. Finally, we computed the sensitivity, specificity, ROC and AUC\footnote{We are aware of the issues related to the selection of a \emph{causal} model based on its classification performances, hence, we took into account both in-sample and out-of-sample metrics.} for each CBN model.

\begin{definition}[Sensitivity \& Specificity]
    Given a binary classification problem, the confusion matrix is a $2 \times 2$ squared integer matrix resulting from the application of a classification algorithm. The values on the main diagonal are called true positives ($TP$) and true negatives ($TN$), while the values on the off diagonal are false positives ($FP$) and false negatives ($FN$). Then, the true positive ratio ($TPR$) and the true negative ratio ($TNR$) are defined as follows: \\

    \begin{minipage}{.4\textwidth}
      \begin{equation}
          TPR = \frac{TP}{TP + FN}
      \end{equation}
      \\
    \end{minipage}% This must go next to `\end{minipage}`
    \begin{minipage}{.4\textwidth}
      \begin{equation}
          TPR = \frac{TN}{TN + FP}
      \end{equation}
      \\
    \end{minipage}
    
    \noindent The $TPR$ and $TNR$ are also called sensitivity and specificity, respectively.
\end{definition}

\begin{definition}[ROC \& AUC]
    The Receiving Operating Characteristic (ROC) curve is a plot of sensitivity and (1 - specificity) measures at different thresholds. The Area Under the Curve (AUC) is the area under the ROC curve.
\label{def:auc-roc}
\end{definition}

\begin{algorithm}
\caption{Confidence matrix from missing data and prior knowledge.}
\label{alg:confidence}
\begin{algorithmic}[1]
\Procedure{ConfidenceMatrix}{$\mathbf{D}, \mathbf{K}, \pmb{\alpha}, n, m$}
    \State $\mathbf{C} \gets \mathbf{0}$ \Comment{Initialize a $|\mathbf{V}| \times |\mathbf{V}|$ matrix, with $\mathbf{V}$ the variables in $\mathbf{D}$.}
    \For{$i \in [1, n]$} % \subset \mathbb{N}$} \Comment{}
        \State $\mathbf{D}_i \gets \Call{Sample}{\mathbf{D}, m}$ \Comment{Sample from $\mathbf{D}$ with replacement.}
        \State $\mathcal{G}_i \gets \Call{StructuralEM}{\mathbf{D}_i, \mathbf{K}, \pmb{\alpha}}$ \Comment{Learn $\mathcal{G}_i$ from $\mathbf{D}_i$ and $\mathbf{K}$.}
        \State $\mathbf{C}[X, Y] \gets \mathbf{C}[X, Y] + 1, \quad \forall \, (X, Y) \in \mathbf{E}_i$ \Comment{Increment the edge count.}
    \EndFor
    \State $\mathbf{C} \gets \mathbf{C} / n$ \Comment{Normalize the confidence matrix.}
    \State \Return $\mathbf{C}$
\EndProcedure
\end{algorithmic}
\end{algorithm}
\begin{algorithm}
\caption{Learn CBN from missing data and prior knowledge.}
\label{alg:learncbn}
\begin{algorithmic}[1]
\Procedure{CBN}{$\mathbf{D}, \mathbf{K}, \pmb{\alpha}, n, m, \lambda$}
    \State $\mathcal{G} \gets (\mathbf{V}, \varnothing)$ \Comment{Initialize an empty graph over the variables $\mathbf{V}$ in $\mathbf{D}$.}
    \State $\mathbf{C} \gets \Call{ConfidenceMatrix}{\mathbf{D}, \mathbf{K}, \pmb{\alpha}, n, m}$ \Comment{Compute the confidence matrix.}
    % \State $\mathcal{G} \gets G \cup (X, Y), \quad \forall \, (X, Y) \quad s.t. \quad \mathbf{C}[X, Y] > \lambda$
    \State \textit{Insert edges into $\mathcal{G}$ following a strategy w.r.t. $\mathbf{C}$ and $\lambda$.}
    \State $\hat{\Theta} \gets \Call{EM}{\mathcal{G}, \mathbf{D}}$ \Comment{Estimate the parameters using EM.}
    \State $\mathcal{B} \gets (\mathcal{G}, \hat{\Theta})$ \Comment{Build the CBN given $\mathcal{G}$ and $\hat{\Theta}$.}
    \State \Return $\mathcal{B}$
\EndProcedure
\end{algorithmic}
\end{algorithm}
\begin{figure}
    \centering
    \includegraphics[scale=0.70]{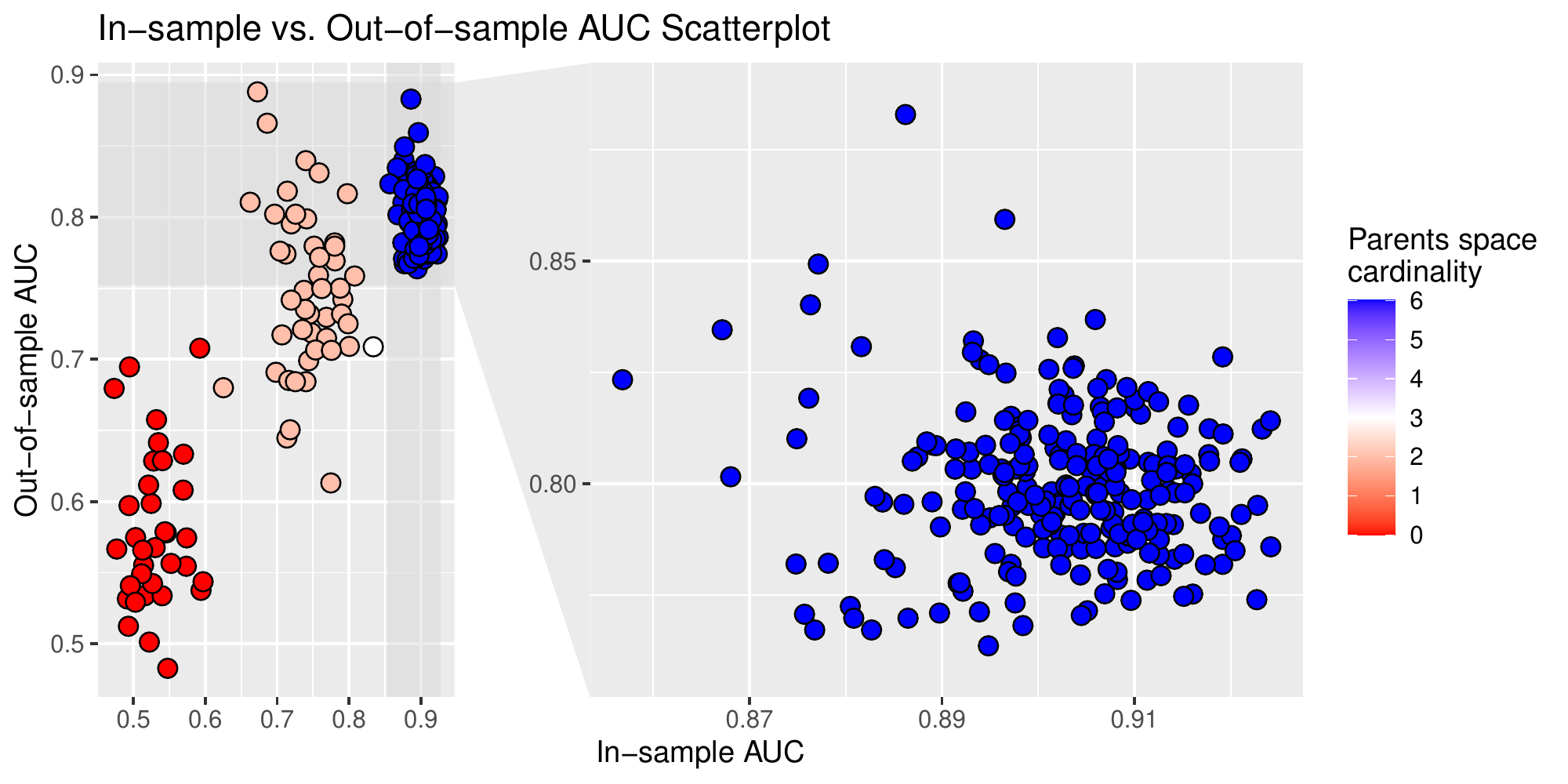}
    \caption{Scatter plot of the results of \Cref{alg:learncbn}. Each dot is a CBN with achieved in-sample and out-of-sample AUC on the horizontal and vertical axes, respectively. Dots color depend on the cardinality of the space of the parents of the target node LNM. To the right, a zoom of the cluster of those CBN that achieved higher values of AUC.}
    \label{fig:auc_parents}
\end{figure}

\noindent \Cref{fig:auc_parents} is a scatter plot of the results of the execution of \Cref{alg:learncbn}. The color mapping allows to clearly distinguish three well-separated clusters, grouped by the parents space cardinality. Specifically, the red cluster represents the models where \textit{LNM} has no parents, the light-red cluster contains models where \textit{LNM} has only \textit{Chemotherapy} as parent, and finally the blue cluster where both \textit{Chemotherapy} and \textit{Histology} are parents of \textit{LNM}.

\begin{figure}
    \centering
    \includegraphics[scale=1.10]{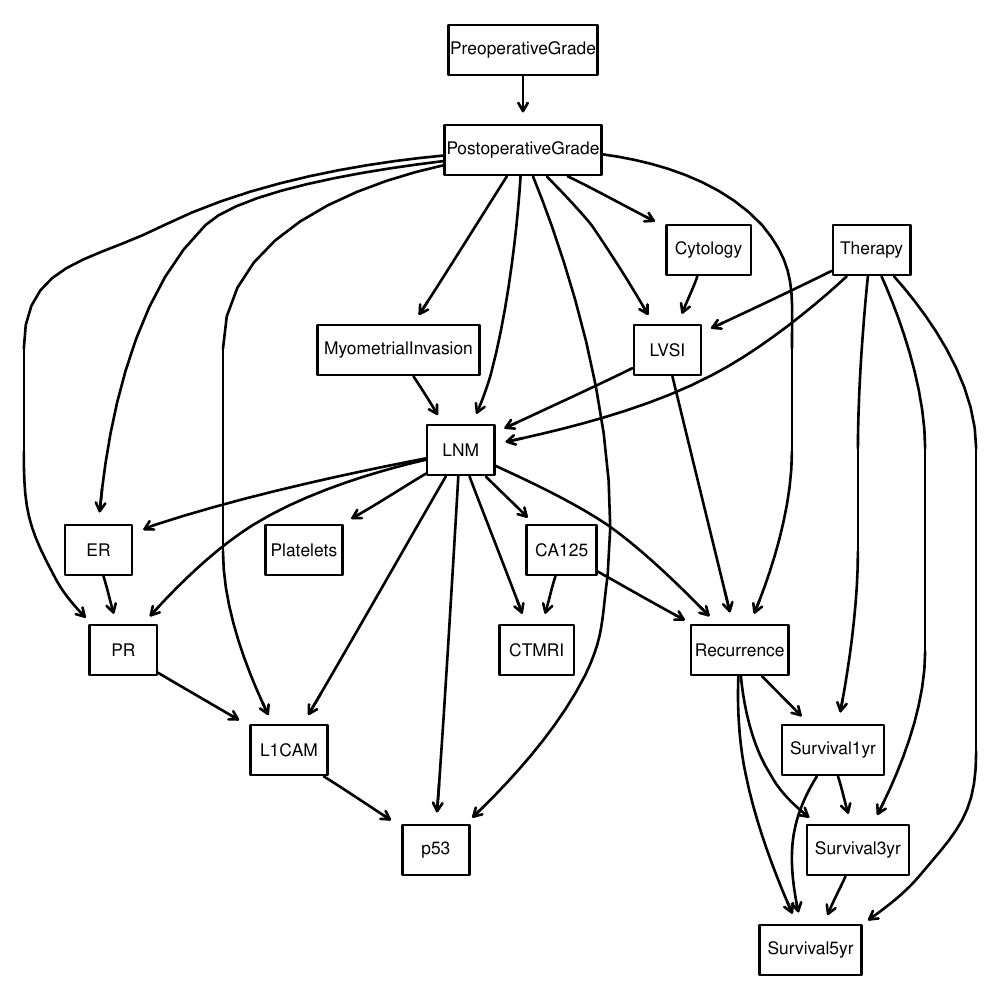}
    \caption{Reference graph obtained by experts' knowledge and randomized controlled trial.}
    \label{fig:reference}
\end{figure}

The structure presented in \Cref{fig:reference} is built by encoding prior causal knowledge elicited by clinicians and randomized controlled trials (RCTs). The encoding process is performed by adding a directed edge from the expected cause to its effect. Each edge addition is supported by biological and physiological knowledge, either obtained by querying experts or from reviewed literature, without observational data. Note, for example, that the \textit{Therapy} node does not have incoming edges, since therapy is always assigned at random in an RCT (and only the outcome matters).

\begin{figure}
    \centering
    \includegraphics[scale=0.70]{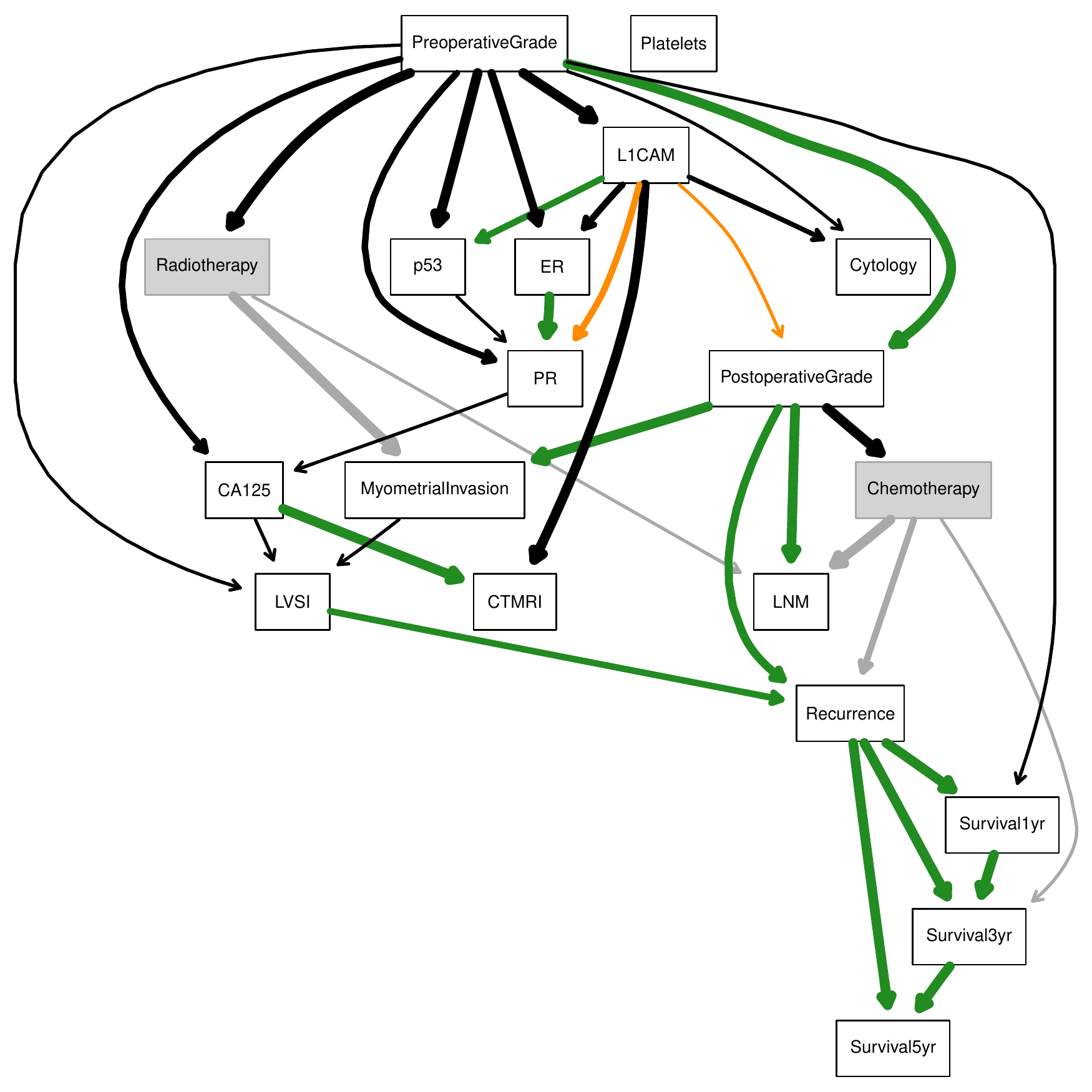}
    \caption{Strength plot of recovered CBN. The edges thickness depends on the strength of the edge itself, which is estimated by the confidence matrix $\mathbf{C}$. The nodes and edges are colored to ease the comparison with the reference graph \Cref{fig:reference}. In particular, green edges are present both in reference and recovered graph with the same orientation, orange edges are present both in reference and recovered graph with reversed orientation, gray nodes and edges cannot be directly compared due to different node sets, and, finally, black edges are present only in the recovered graph.}
    \label{fig:strength_no-hospital_tiers}
\end{figure}

The graph presented in \Cref{fig:strength_no-hospital_tiers} is the result of the application of \Cref{alg:learncbn} on the collected data set and encoded prior knowledge based on partial temporal ordering of variables. The \textit{Therapy} node is split into \textit{Radiotherapy} and \textit{Chemotherapy} to highlight the different impact of adjuvant treatments.

\begin{figure}
    \centering
    \includegraphics[scale=0.70]{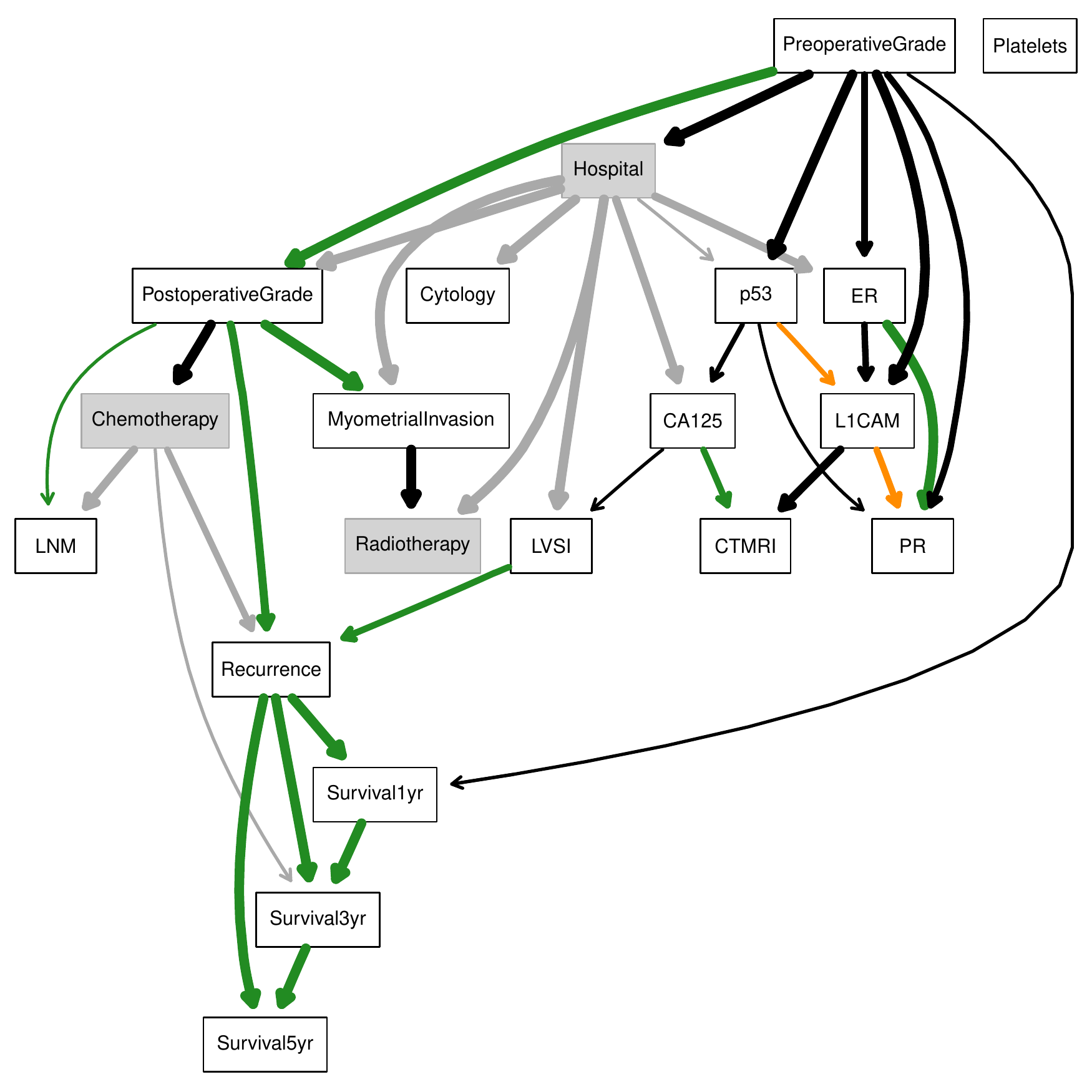}
    \caption{Strength plot of recovered CBN with the addition of the previously unobserved Hospital variable. The edges thickness and the coloring schema are the same of \Cref{fig:strength_no-hospital_tiers}.}
    \label{fig:strength_hospital_tiers}
\end{figure}

The two graphs share a common subset of edges, e.g. the ones related to \textit{Recurrence} and \textit{Survivals}. A major difference stands in the edges related to the biomarkers cluster. Indeed, while in \Cref{fig:reference} biomarkers, such as \textit{p53}, \textit{CA125} and \textit{L1CAM}, are assumed to be strongly related to \textit{LNM}, in the recovered graph the \textit{PreoperativeGrade} is observed as common parent of the variables contained in such cluster. Moreover, no biomarker is directly connected to \textit{LNM}, not as a parent nor as a child, calling for further analyses of the collected data.

Such similarities and differences also appear in \Cref{fig:strength_hospital_tiers}, where \textit{Hospital} is introduced to explore the potential presence of latent effects and selection bias. While the graph in \Cref{fig:strength_hospital_tiers} is not completely different from the one in \Cref{fig:strength_no-hospital_tiers} in terms of observed substructures, the latter encodes different independence statements due to the presence of the newly introduced \textit{Hospital}.

The crucial difference stands in the semantic interpretation of Hospital, which in this case is not to be intended as a direct cause of its children, but rather as a proxy for others unobserved variables or biases, i.e. a context variable. Indeed, while it could be that population heterogeneity across hospitals affects the choice of adjuvant treatments, it would be nonsensical to conclude that Hospital is a cause of Ca-125. Nonetheless, the causal discovery procedure includes a set of edges that are related to spurious associations present in the data set. For example, the directed edge that connects Hospital to p53 is an instance of such pattern, which could be caused by a missing-not-at-random (MNAR) mechanism \cite{Scutari2020BayesianData}. Another example of the impact of biases is represented by the directed edge from Hospital to PostoperativeGrade. In this case, an unbalanced distribution of patients' grading across geographical regions, which Hospital is a proxy of, could act as a potential source of selection bias \cite{Esterling2021TheClaims}.

\begin{figure}
    \centering
    \includegraphics[scale=0.50]{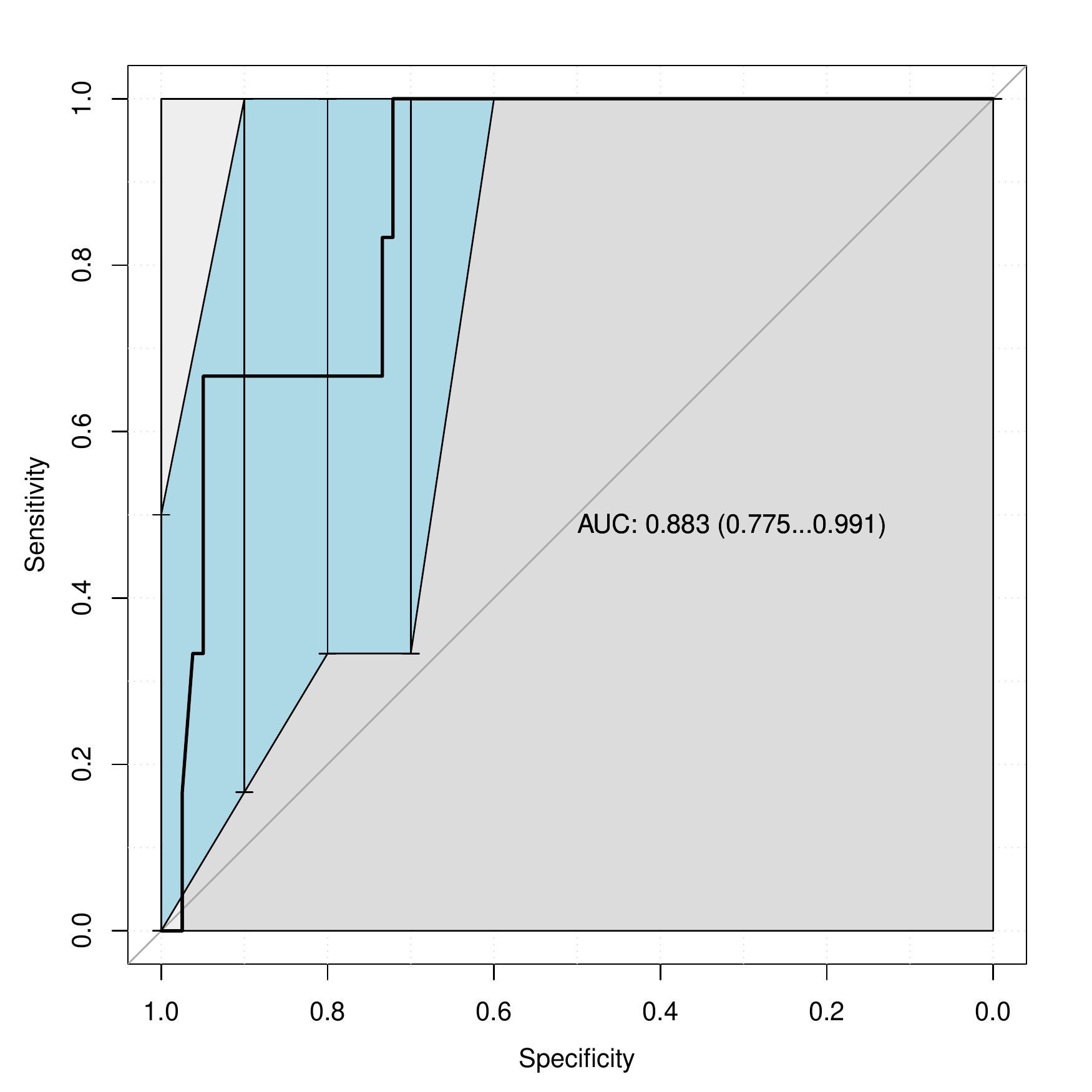}
    \caption{ROC curve and AUC value of the CBN model fitted from the graph in \Cref{fig:strength_hospital_tiers} and train data.}
    \label{fig:roc_hospital_tiers}
\end{figure}

The ROC curve depicted in \Cref{fig:roc_hospital_tiers} is obtained by predicting the probability of the LNM class on the test set, given the CBN fitted on the structure in \Cref{fig:strength_hospital_tiers} and the train set. It achieves an AUC of 0.883, with associated 95\% CI 0.775-0.991, which is higher than the one obtained in \cite{Reijnen2020PreoperativeStudy}, although it was not possible to compare the metrics using a significance test due to the different test sets.

\section{Conclusions and Future Works}

Given the known limitations of data-driven approaches when applied to observational data, causal discovery techniques are used to explore and mitigate the impact of spurious associations during the learning process. In this work we explored the task of learning a causal representation to assess the pre-operative risk of developing LNM in endometrial cancer patients. Furthermore, the recovered models were extended to include information from context variables, aiming to uncover previously unobserved effects.

The resulting procedure takes advantage of pre-existing techniques to reduce the bias introduced during the imputation step in a bootstrap approach. This enabled us to compute the strength of the observed associations in the obtained models across multiple re-sampled instances, allowing a step of model averaging to recover less frequent substructures. The risk assessment is performed by predicting the probability of developing LNM using a CBN fitted on the recovered structure and given train set, showing an increased AUC over previous works.

Still, we highlighted a set of potential issues that need to be addressed in future works.

\paragraph{Missingness Mechanism.} With the introduction of the Hospital variable we observed a set of edges that hint to the presence of a potential missing-not-at-random pattern. If this is the case, then it would require careful consideration in order to reduce the bias introduced during the missing imputation step.

\paragraph{Effect of Adjuvant Therapy.} Once a causal graph is obtained, it is theoretically possible to estimate the causal effect of each adjuvant therapy, either single or combined, on the development of LNM. Before directly computing the the effect, there are assumptions that need to be carefully verified, e.g. positivity, consistency, unconfoundedness and non interference \cite{pearl2016causal}.

\paragraph{Impact of Selection Bias.} While it is clear that observing an association between Hospital and other variables it is not sufficient to conclude that, indeed, there is a selection bias, it is a strong hint that there are other unobserved variables that influence the causal mechanism. It could be interesting to assess which is the impact of the selection bias mediated by the Hospital variable alone.

% \section*{Funding} \small{Alessio Zanga was granted a Ph.D. scholarship by F. Hoffmann-La Roche Ltd.}
\section*{Acknowledgments}
Alessio Zanga was granted a Ph.D. scholarship by F. Hoffmann-La Roche Ltd.

\bibliography{references}
\bibliographystyle{plain}

\end{document}